# Deep Belief Networks Based Feature Generation and Regression for Predicting Wind Power

Asifullah Khan, Aneela Zameer*, Tauseef Jamal, Ahmad Raza

Email Addresses: ([asif, aneelaz, jamal]@pieas.edu.pk, raza.aflak@gmail.com)

Department of Computer Science, Pakistan Institute of Engineering and Applied Sciences (PIEAS), Nilore, Islamabad, Pakistan.

Aneela Zameer*:

Email: aneelaz@pieas.edu.pk; aneelas@gmail.com

Phone: +92 3219799379

Fax: 092 519248600

## ABSTRACT

Wind energy forecasting helps to manage power production, and hence, reduces energy cost. Deep Neural Networks (DNN) mimics hierarchical learning in the human brain and thus possesses hierarchical, distributed, and multi-task learning capabilities. Based on aforementioned characteristics, we report Deep Belief Network (DBN) based forecast engine for wind power prediction because of its good generalization and unsupervised pre-training attributes. The proposed DBN-WP forecast engine, which exhibits stochastic feature generation capabilities and is composed of multiple Restricted Boltzmann Machines, generates suitable features for wind power prediction using atmospheric properties as input. DBN-WP, due to its unsupervised pre-training of RBM layers and generalization capabilities, is able to learn the fluctuations in the meteorological properties and thus is able to perform effective mapping of the wind power. In the deep network, a regression layer is appended at the end to predict sort-term wind power. It is experimentally shown that the deep learning and unsupervised pre-training capabilities of DBN based model has comparable and in some cases better results than hybrid and complex learning techniques proposed for wind power prediction. The proposed prediction system based on DBN, achieves mean values of RMSE, MAE and SDE as 0.124, 0.083 and 0.122, respectively. Statistical analysis of several independent executions of the proposed DBN-WP wind power prediction system demonstrates the stability of the system. The proposed DBN-WP architecture is easy to implement and offers generalization as regards the change in location of the wind farm is concerned.

# 1. INTRODUCTION

One of the key challenges now a day is to find alternate energy sources with the well-known fact that energy sources like coal, oil, and gas will deplete with the passage of time; even water and nuclear fuel is available in limited amount. Consequently, industries are looking forward for researchers to explore efficient and reliable methods to exploit unlimited energy resources such as wind and sun, and production of energy on commercial scale. Complex processes, such as oil and gas extraction and refining, are not required for these alternate sources of energy and therefore, these unlimited sources are comparatively cheaper. Moreover, wind power generation has the capability of producing power on large scale and it is also pollution free, as well as being renewable source of energy, there will not be a question for future wind supply. Therefore, there is a trend to switch on to wind energy on massive scale in the developing countries; many European countries have already started exploiting wind power generation. Similarly, America and Canada are also utilizing heavy wind density to generate wind energy. However, accurate wind power prediction is a characteristic feature for wind power generation and integration of the power plant.

The primary issue with wind power generation is to provide steady power distribution. Since wind is fluctuating source of energy, power produced by wind at any instance is not known in advance. Therefore, in order to achieve smooth power distribution, prediction of wind power in advance for short or long time duration is required. Forecast of the wind power for short time duration spams over a range that vary from minutes to a day, while long-term forecast on the other hand varies from days to months and years. It is a known fact that the generated wind power depends largely upon the speed of wind. There are two different types of models which are used for prediction of wind power namely statistical and physical [1-3]. Physical models are based on physical laws that govern the atmospheric behavior. Different Parameters such as roughness of surface, pressure, temperature, humidity, obstacles of atmosphere and orography are evaluated to forecast wind power and speed. Statistical approaches are also evaluated for the same problem and are based on wind's stochastic nature, i.e. a relationship between wind power and factors associated with it based on history [4]. In our work, we used a machine learning based statistical approach to learn from historical data and forecast short-term wind power for steady power distribution.

ANNs are empowered with ability to learn complex mappings. One of the initial works related to use of ANNs in wind power forecasting was gust prediction carried out by Carcangiu et al. [5]. S. Li et al. developed a wind forecasting model by devising new feature selection technique for selection of compact set of input features [6]. While, G. Grassi et al. predicted the wind energy using two hidden layers [7]. In [8], deep neural networks based ultra-short-term wind forecast model is suggested. Jursa et al. solved short-term wind power forecast problem using artificial intelligence [9]. They exploited evolutionary algorithms to reduce prediction error. Jursa's model used nearest neighbor search along with ANNs to forecast wind power on hourly basis. While Grassi's model was capable of predicting wind power on monthly basis and used ANNs [7]. García et al. [10] used series of regression Support Vector Machines for devising prediction system for short-term wind speed.

A series of hybrid approaches has shown significant improvement based on two or more techniques [11-17]. Among these, Amjady et al. proposed interesting but a bit complex model to improve prediction. Amjady model used irrelevancy and redundancy filters for feature selection, which are then passed to forecast engine [11]. Forecast engine contains distinctive neural networks along with Enhance Particle Swarm Optimization for weight adjustment. Neural networks used in their forecast engine are: Levenberg-Marquardt, Bayesian Regularization and Broyden-Fletcher-Goldfarb-Shanno.

In [12], the authors used a two stage hybrid approach for feature selection for price spike forecast of electricity markets. The hybrid approach proposed in [13] is composed of neural networks and evolutionary algorithm. Andrew et al. [14] analyzed the capabilities of data mining approaches for wind power perdition of a wind form. On the other hand, A. Zameer et al. [18] proposed ensemble based technique, GPeANNs in which multiple artificial neural networks are combined and performance of model is optimized using genetic programming algorithm (GP). GPeANNs is a two phase wind power prediction model, which uses 70% of available data for training of five neural networks in first phase. Whereas, in the second phase it makes use of 30% of unseen data separated from training data and forecasts that are obtained from neural networks. In this phase, GP is utilized for tree generation and for defining a mathematical relationship between input features and target vectors. Hybrid technique comprising variational model decomposition with a form of extreme learning machine has also been used for short-term speed prediction [19]. Overall, an in-depth review of various statistical and hybrid approaches can be found in the literature [20-24].

Statistical models are also incorporated in hybrid form for wind speed or wind power prediction. In this regard, recently used least square support vector machine optimized with bat algorithm for wind power prediction and a self-adaptive ARIMAX Model with an Exogenous WRF Simulation for wind speed prediction have been reported [25, 26]. Capabilities of machine learning strategies have been intensively used in diverse areas of Science and Engineering such as prediction in materials science [27], knowledge acquisition [28], EEG and brain connectivity [29,30], and other industrial applications [31]. Similarly, machine learning has been employed to intelligently perform watermarking of an image [32-33]. Specifically, the utilization of learning capabilities of ANNs can be found in various fields [34-38].

Recently, several interesting machine learning techniques have been employed for the development of short-term wind power predication models. However, it is observed that models with good prediction power are compromised with high complexity in order to increase the prediction accuracy, the reported methods are complex (of hybrid nature using several types of regressors/learners, different feature extraction, and reduction techniques). Utilizing the strength of deep neural networks, many researchers have reported their work on wind power and wind speed predictions in [39-42]. In this work, we show that deep learning mechanism such as DBN can be effectively used to solve the wind power prediction problem. The proposed DBN based Wind Power prediction system (DBN-WP), due to its deep hierarchical learning and good generalization capabilities, not only yields improved wind power prediction but also keeps the prediction system simple and easy to implement by avoiding complex and hybrid machine learning mechanisms. Feature selection capability due to of pre-training makes it simple to use. As contrast to traditional ANN approaches applied to wind power or speed prediction, the proposed model makes use of recent advances in deep learning technique, for instance, initializing parameter via unsupervised pre-training of DBN, hence enhancing the prediction accuracy.

## 2. METHODS

### 2.1. DEEP NEURAL NETWORK BASED LEARNING

Deep learning networks are learning mechanism, where high-level abstractions in data are modeled by using multiple processing layers. Therefore, deep learning techniques are being mostly favored for complex problems of both unsupervised and semi-supervised nature. Deep belief network (DBN) is a learning mechanism based on deep neural network having capability of unsupervised pre-learning. DBN is composed of multiple Restricted Boltzmann Machines (RBMs). DBN first perform layer-wise greedy and unsupervised learning, whereby each layer is represented by an RBM. DBN is trained in an unsupervised way (without target labels) to reconstruct its inputs. Each layer of DBN acts as feature generator and converts the input to more abstract representation. After unsupervised learning, DBN can further be fine-tuned through supervised learning (use target labels) to perform classification or regression. This supervised learning after the unsupervised greedy layered-wise training is mostly performed using gradient descent. Approach of deep neural networks has been used in wide applications such as devising predictive feature space for object detection in natural images and in medical diagnostics [43-44].

### 2.1.1. Restricted Boltzmann Machines

RBM [15] is a two layer generative neural network which is built upon probabilistic binary units that works in stochastic manner. One layer is called visible layer and denoted by v, while the other layer is called hidden layer and denoted by h. Every unit in one layer is connected to all units of other layer and these connections are bi-directional. No connection between units of same layer exists. Neurons of RBM form a bipartite graph and therefore, it is called Restricted Boltzmann Machine in contrast to Boltzmann Machine [45, 46].

RBM is an energy based model in which each configuration of the variables of the system is associated with energy value. Configurations with lower energy are more probable. Energy of a given configuration is defined in equation 1, where W is the weight matrix of connections and b are c are used to define the bias vector of visible and hidden layer respectively. The probability of each configuration is inversely proportional to the energy of variables configurations, as given by equation 2.

$$E(v,h) = v^T W h + b^T v + c^T h \quad (1)$$

$$P(v,h) = \frac{1}{z} e^{-E(v,h)} \quad (2)$$

where Z is used to define the partition factor and is defined as:

$$Z = \sum_{v,h} e^{-E(v,h)} \quad (3)$$

Since no connections exist between units of same layer, therefore the activation of the each of the hidden unit is mutually independent for the given visible units activation and same is true for visible layer. Probability of units in each layer depends on the state of the other layer and is computed as in equation 4-6.

$$P(v_j = 1 | h) = \Omega(-b_j - W_j h) \tag{4}$$

$$P(h_j = 1 | v) = \Omega(-c_j - W_j^T h) \tag{5}$$

Where
$$\Omega(x) = \frac{1}{1 + e^{-x}} \tag{6}$$

Training of RBM is usually performed through Contrastive Divergence (CD) algorithm [45]. The CD algorithm is based on Gibbs sampling and is used for weight matrix updation inside the gradient descent procedure. Following are the steps of CD algorithm for a single sample:

- For each training example $v$, the probability of hidden unit is computed and after sampling, hidden activation vector is prepared.
- Compute the positive gradient denoted by product $vh^T$.
- Hidden layer activation $h$ computed in the previous step is used to sample a visible activation vector $v'$ on visible layer and this is used again to sample hidden activation vector $h'$.
- Compute the negative gradient denoted by product $v'h'^T$. The weight update is then given by

$\Delta W_{ij} = \alpha(vh^T - v'h'^T)$, where α is learning rate.

### 2.1.2. Deep Belief Network

DBN is a stack of many RBMs such that hidden layer of each RBM act as visible layer for next RBM. Visible layer of first RBM is also visible layer of the DBN and all other layers are hidden layers of DBN. DBN is trained by training a single RBM at a time. Once, first RBM has been trained, training samples are simply forwarded through it and the output produced at hidden layer of it, is served as input on the visible layer of next RBM and so on. This is called layer-wise pre-training of DBN. Figure 1 shows the basic architecture of DBN used in the proposed DBN-WP.

The layer-wise greedy and unsupervised pre-training of DBN allows us to train a network with a large number of hidden layers; this may not be possible with back propagation algorithm [47]. Once all the RBMs in the DBN have been trained, we can simply convert the bi-directional weights to unidirectional weights and DBN is ready to be used as a neural network. The benefit of converting DBN to NN is that we get an NN having large number of hidden layers, which are already pre-trained to some extent. To convert this NN to predict labels associated with training samples, a layer at the top is added that in which number of neurons are equal to number of classes. Now this whole NN is finally trained with back-propagation method for fine tuning. Since weights have already been initialized already with pre-training, training of this NN is accomplished much faster and in a more efficient way.

### 2.1.3. DBN used as a Regressor for Prediction

To convert a DBN to a regressor, there exist two possibilities [16]. First method is to train DBN in an unsupervised way and then convert it to neural network without adding any other layer. Now all the samples are fed forward through this NN and the output is treated as samples for regression model. The later approach is to train DBN via unsupervised learning and convert it to neural network by adding an outer layer (usually with one neuron) for regression on the top. The neural network is then trained with labeled data for fine tuning and used as regression model. In this work, we have employed the later approach.

### 2.2. PROPOSED DBN-WP BASED WIND POWER PREDICTION SYSTEM

It is well known that DBN represents a valuable tool for learning. Our proposed DBN-WP technique is used for performing both feature generation and regression using DBN. Our DBN architecture contains several (3-4) hidden layers comprising of RBMs. We have employed two architectures for the proposed DBN-WP system, denoted as DBN-WP1 and DBN-WP2. RBMs work on principal of minimizing free energy, while performing layer-wise pre-training.

The proposed DBN-WP exploits meteorological properties (wind speed, pressure, temperature etc.) to predict the wind power. DBN-WP, due to its unsupervised pre-training of RBM layers and generalization capabilities, is able to learn the fluctuations in the meteorological properties and thus is able to perform effective mapping of the wind power. The proposed DBN-WP based prediction consequently, is capable of not only providing better power prediction than existing techniques but, also shows good generalization by performing better on unseen data. Furthermore, overall structure of the proposed model is much simpler in contrast to the existing complex models that contain multiple neural networks and other machine learning elements, such as Amjady's model [11]. Figure 2 shows the basic architecture of the DBN, while Figure 3(a) shows the proposed DBN-WP model in which DBN has been converted to a regressor by adding one neuron at top.

### 2.2.1. Dataset

Performance of the proposed wind power prediction system, DBN-WP is evaluated on standard data sets [31]. The data sets are comprised of three years' data from five (05) wind farms of Europe having the similar geographical region. The feature vector of dataset includes both historical measurements and meteorological predictions for wind components. Historical power measurements are available with intermediate (hourly) temporal resolution. In order to exploit features of all wind farms and to facilitate scale independent experimentation, the feature vector has been normalized between 0-1.

Meteorological forecasts as well as zonal ($K_z$) and meridional ($K_m$) components of wind speed and direction were collected at 10 meters above the surface. Medium-range Weather Forecasts has been used for the collection of aforementioned predictions. In this deterministic predictions are used that are released twice per day. Each prediction is collected at duration of 1 hour and 2 days afterwards. Four weather predictions are collected for each power value that are released from past 48 hours and each is separated by 12 hours.

Basically four weather predictions are acquired corresponding to each value of measured power; each composed of direction, speed, zonal and meridional components. These predictions have been released at a time scale with steps of 12 each, i.e. '*t*', '*t+12*', '*t+24*' and '*t+36*' hours back, but provide hourly forecasts, where *t* is the current time of prediction. Production Horizon is t+1 hour as the present proposal focuses on hourly based power forecast. Input features are comprised of these weather predictions and power measurement values of preceding 24 hours along with the present forecasts. Overall features set *F(t)*, can be expressed mathematically as following equation (7):

$$F(t) = \begin{bmatrix} K_s(t), K_s(t-1), ..., K_s(t-24), \\ K_D(t), K_D(t-1), ..., K_D(t-24), \\ Kz(t), Kz(t-1), ..., Kz(t-24), \\ Km(t), Km(t-1), ..., Km(t-24), \\ K_P(t-1), ..., K_P(t-24) \end{bmatrix} \quad (7)$$

where $K_P(t)$, $K_D(t)$, and $K_s(t)$ denotes power, direction, and speed of wind (at time *t*; where time is in hours), respectively, and $Kz(t)$ and $Km(t)$ represent the zonal- and meridional- components of surface wind as per specification, respectively, again at current time 't'. This feature set goes into the DBN based regressor as input shown in Figure 3(a).

Figure 3(b) illustrates the details of the independent variables in comparison with dependent variable (predicted Wind Power) in the form of block-diagram. The overall input features of the proposed DBN-WP model are 124 in number composed of 25 hourly-values each of U, V, $K_S$, $K_D$, and 24 hourly values of previous wind power $K_P$. All the feature values that are exploited are taken from the forecast dataset. Similar approaches of using weather forecast dataset have already been implemented by many researchers [48-50].

### 2.2.2. DBN-WP Cross-validation using 5-fold Test

Two different architectures for the proposed DBN-WP have empirically selected composed of RBMs. In order to test their performance, both 5-fold and hold-out tests are carried out. The 5-fold test is considered as a stringent test compared to hold-out or self-consistency test [35-36]. In 5-fold test, 5-folds are randomly generated; one is used for testing, while the rest of four are used in training. As 5 fold is used therefore each step is performed five times so that every fold gets a chance in the set under test. Then the performance of proposed wind power predictor is evaluated on dataset by computing difference between predicted and measured values of wind power.

### 2.2.3. DBN-WP Cross-validation using Hold-out Method

We have also evaluated performance of the DBN-WP models using hold-out data processing method. Data is partitioned into training and validation set by specifying 70% for training while assigning rest of the 30% unseen data for validation. Performance is measured by computing the difference between actual and predicted labels.

### 2.2.4. Parameter Setting of the Proposed Technique

We empirically set the parameters of DBN-WP model. Multiple experiments were run to find out the best parameters for which DBN-WP model yields best results. The prominent parameters that can affect accuracy of prediction are depth of network, number of processing units within each layer, activation functions, learning rate, batch-size, and momentum. Table 1 demonstrates the details of parameter settings that yielded best prediction results. Momentum adds a part of the preceding weight to the present one. In essence, it is used to avoid algorithm from congregating to local minima. High value of momentum helps in increasing the speed of convergence of the system at cost of overrunning the minima. On the other hand, too low value of momentum may not consistently avoid local minima, and additionally, can make the training sluggish. Learning

Rate is a parameter that dictates the size of changes in weights and biases during learning of the training algorithm.

## 3. RESULTS

This section describes how our proposed approach is useful in predicting wind energy. The DBN-WP is trained and tested using five standard wind power datasets. We have separately trained and testes our model on dataset of each wind farm. We exploited year 2007-2010 wind farm data were for training of the DBN-WP model, while year 2011 data was used for testing.

### 3.1. Performance Metrics

To evaluate our model in terms of prediction performance, different error-measures (Standard Deviation Error (SDE), Mean Absolute Error (MAE) and Root Mean Square Error (RMSE)) are used as performance indicators. RMSE is estimated through the given equation:

$$RMSE = \sqrt{\frac{1}{N}(y_{actual} - y_{predicted})^2} \qquad (8)$$

where, $y_{actual}$ and $y_{predicted}$ is the actual and predicted wind power measurements respectively for the trained model.

Similarly, MAE is expressed as:

$$MAE = \frac{1}{N}\sum_{i=1}^{N}|y_{actual} - y_{predicted}| \qquad (9)$$

Similarly, SDE can be calculated as:

$$SDE = \sqrt{\frac{1}{N}\sum_{i=1}^{N}(error_i - error_{mean})^2} \qquad (10)$$

where $error = y_{actual} - y_{predicted}$; and $error_{mean}$ is the average value of error. All error measures have units of wind power which is normalized in our case.

### 3.2. Power Analysis for Actual and Predicted Values

The performance of the proposed technique is analyzed for measured wind power and predicted wind power, although input feature set is taken from forecast data set. This analysis is made using two different data processing techniques; hold-out test and 5-fold test. In hold-out test, model is trained on two third of data while one third is used for testing. Therefore, the classifier is trained on two third of the data and the rest of the unseen samples are used as test data. Error measures for predicted wind power, obtained using hold-out data processing are plotted in Figure 4. The hold-out data processing results are compared with that of the 5-fold data processing.

In our proposed model, DBN-WP, 5-fold data processing has also been used. Predicted and actual wind power has been plotted in Figures 5-9 for five wind farms. Only first 300 values are presented for better comparison and visual impact. It can be observed that these graphs represent a good match between predicted and actual wind power curves. Predicted data almost overlaps the actual data for the first three wind farms, which shows that the same model caters different farms and gives overwhelming results with much lower prediction errors. However, a slight difference between the two curves could be examined in case of wind farms 4 and 5, which could be the effect of those physical parameters including specific terrain or any other physical phenomenon whose variables do not contribute much in our model. It can be easily handled separately if the difference would be much larger, but in our case, it produces quite reasonable results. The performance of proposed model DBN-WP is further carried out in terms of error measures. Error measures for 5-folds are presented in Table 2. While Table 3 represents a comparison of mean values of error measures from 5-fold and hold-out. It can be observed (c.f. Figure 4) that mostly 5-fold data processing produced better results.

### 3.3. Performance Comparisons with Existing Schemes

As a next step, we have compared the performance of our proposed technique with base line regressor, Autoregressive Integrated Moving Average (ARIMA), and Amjady et al. [11] and Grassi et al. [7] technique used for wind power prediction. It is important to note here that we have implemented their models on our data set for all five wind farms, trained, tested and then made the comparison. This comparison is made in terms of error measures on the dataset, and the resulted values are illustrated in Table 4. Pictorially it is demonstrated in Figure 10. Examining this figure, it can be concluded that almost all values of the error measures leads to smaller values with the proposed DBN-WP model. Furthermore, the proposed model is better in the sense that it does not make use of hybrid schemes and do not need to go through all feature selection/extraction phases. It further shows the robustness of our model. It has been observed that the deep learning mechanism of DBN supports in predicting short-term wind power with more accuracy and more efficiently.

### 3.4. Computational Time Based Comparisons

Since the proposed technique does not use any explicit feature selection technique, therefore, it was expected that it would take less time in both training and testing phases. For instance, computational time using hold-out data processing for wind farm-5 is shown in Table 5.

### 3.5. Statistical Analysis of DBN-WP using Several Runs

In order to provide statistical analysis of the prediction performance of the proposed technique, several independent executions of DBN-WP were made. The code was executed for 100 independent runs and different statistical measures were computed and sorted, as shown in Figure 11. It can be observed from these curves that the proposed solution is stable with the variations of around 2% for RMSE and MAE, and around 1% for SDE from their mean values of 0.124, 0.083 and 0.122, respectively.

## 4. CONCLUSIONS

This work proposes deep learning based wind power prediction system, to improve both feature extraction as well as prediction. We can conclude from this study that the non-linear mapping capabilities and depth in architecture make it possible for DBN to effectively predict wind power.

The proposed DBN-WP architecture is easy to implement and offers generalization. DBN-WP has been applied on five wind farms and it produced effective results which demonstrate the robustness of this model. Furthermore, execution time for our model is less as compared to other hybrid techniques as we do not need to employ separate feature extraction and selection. The proposed DBN-WP algorithm achieves mean values of RMSE, MAE and SDE as 0.124, 0.083 and 0.122, respectively including all wind farms. Statistical analysis of several runs show that the proposed DBN-WP wind power prediction system is stable.

**Compliance with Ethical Standards**

All the authors of the manuscript declared that there is no

- Potential conflicts of interest
- Research involving Human Participants and/or Animal
- Material that required informed consent

**Competing interests**

All of authors including myself for this submission declare that authors have no financial and non-financial competing interests. Authors declare that there is no conflict of interest.

Table 1 : Parameter Settings of DBNs.

| Type of Deep Learning Networks | Neurons: Hidden Layer | Epochs | Batch size | Learning rate α [range: 0-1] | Momentum [range: 0-1] |
|---|---|---|---|---|---|
| DBN-1 | [100 80 50 5] | 100 | 100 | 0.87 | 0.05 |
| DBN-2 | [80 50 5] | 100 | 100 | 0.90 | 0.05 |

Table 2: Error Measures for the Five Wind Farms using 5-fold Cross-validation.

|  |  | Fold 1 | Fold 2 | Fold 3 | Fold 4 | Fold 5 |
|---|---|---|---|---|---|---|
| MAE | WF 1 | 0.095033 | 0.090100 | 0.092706 | 0.086958 | 0.075899951 |
|  | WF 2 | 0.069278 | 0.071426 | 0.082282 | 0.081351 | 0.072640856 |
|  | WF 3 | 0.095033 | 0.090100 | 0.092706 | 0.086958 | 0.075899951 |
|  | WF 4 | 0.070779 | 0.086325 | 0.073661 | 0.082845 | 0.070825389 |
|  | WF 5 | 0.087518 | 0.084431 | 0.085898 | 0.080906 | 0.085011657 |
| RMSE | WF 1 | 0.125619 | 0.119345 | 0.121937 | 0.118273 | 0.105164558 |
|  | WF 2 | 0.093605 | 0.09629 | 0.108558 | 0.108096 | 0.098419542 |
|  | WF 3 | 0.125619 | 0.119345 | 0.121937 | 0.118273 | 0.105164558 |
|  | WF 4 | 0.097405 | 0.111251 | 0.100165 | 0.108178 | 0.095966 |
|  | WF 5 | 0.122981 | 0.11998 | 0.120797 | 0.110487 | 0.118756184 |
| SDE | WF 1 | 0.116749 | 0.113004 | 0.110105 | 0.117412 | 0.104642495 |
|  | WF 2 | 0.093512 | 0.096046 | 0.108549 | 0.106975 | 0.098396302 |
|  | WF 3 | 0.116749 | 0.113004 | 0.110105 | 0.117412 | 0.104642495 |
|  | WF 4 | 0.094503 | 0.095256 | 0.093758 | 0.092568 | 0.092589598 |
|  | WF 5 | 0.113863 | 0.115556 | 0.116017 | 0.110437 | 0.118616738 |

Table 3: Mean Values of Error Measures on Holdout and 5-folds for all Five Wind Farms.

|   |   | Five-fold | Hold-out |
|---|---|---|---|
| Wind Farm 1 | MAE | 0.0881 | 0.0822 |
|   | RMSE | 0.1428 | 0.1181 |
|   | SDE | 0.1393 | 0.1172 |
| Wind Farm 2 | MAE | 0.0754 | 0.091 |
|   | RMSE | 0.1010 | 0.1212 |
|   | SDE | 0.1007 | 0.1211 |
| Wind Farm 3 | MAE | 0.0881 | 0.0807 |
|   | RMSE | 0.1181 | 0.1225 |
|   | SDE | 0.1124 | 0.1173 |
| Wind Farm 4 | MAE | 0.0769 | 0.0846 |
|   | RMSE | 0.1026 | 0.1132 |
|   | SDE | 0.0937 | 0.1132 |
| Wind Farm 5 | MAE | 0.0848 | 0.0791 |
|   | RMSE | 0.1186 | 0.1214 |
|   | SDE | 0.1149 | 0.1183 |

Table 4: Comparison on the basis of error measures of test data for all five Wind Farms

|  |  | ARIMA Model | Two Layer NN [7] | Modified Hybrid NN [11] | Proposed DBN-WP |
|---|---|---|---|---|---|
| Wind Farm 1 | MAE | 0.4470 | 0.1708 | 0.1100 | **0.0881** |
|  | RMSE | 0.5410 | 0.2215 | 0.1365 | **0.1181** |
|  | SDE | 0.3776 | 0.1860 | 0.1144 | **0.1124** |
| Wind Farm 2 | MAE | 0.4432 | 0.1352 | 0.0879 | **0.0754** |
|  | RMSE | 0.5543 | 0.1792 | 0.1206 | **0.1010** |
|  | SDE | 0.3956 | 0.1738 | 0.1195 | **0.1007** |
| Wind Farm 3 | MAE | 0.5711 | 0.1405 | 0.0972 | **0.0881** |
|  | RMSE | 0.6832 | 0.1888 | 0.1388 | **0.1181** |
|  | SDE | 0.4929 | 0.1821 | 0.1378 | **0.1124** |
| Wind Farm 4 | MAE | 0.4897 | 0.1516 | 0.0867 | **0.0769** |
|  | RMSE | 0.6117 | 0.1963 | 0.1174 | **0.1026** |
|  | SDE | 0.4433 | 0.1703 | 0.1153 | **0.0937** |
| Wind Farm 5 | MAE | 0.5071 | 0.1616 | 0.0872 | **0.0848** |
|  | RMSE | 0.6178 | 0.2124 | 0.1236 | **0.1186** |
|  | SDE | 0.4589 | **0.1020** | 0.1125 | 0.1149 |

Table 5: Computational Time for Wind Farm-5 using Hold-out Data Processing Technique

|  | Training (sec) | Testing (sec) |
|---|---|---|
| DBN-1 | 0.514 | 0.045 |
| DBN-2 | 0.743 | 0.035 |

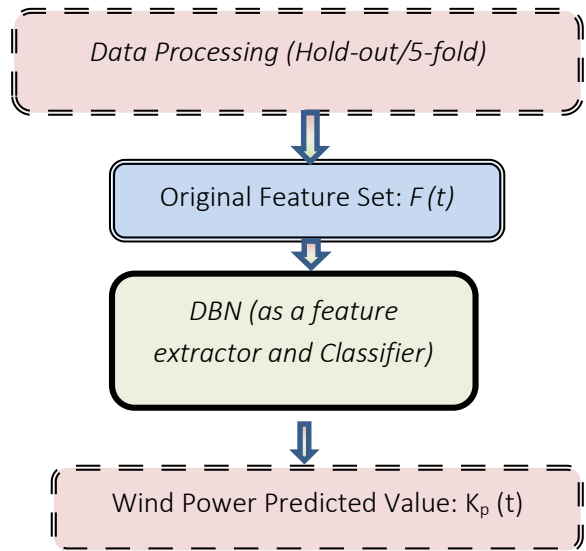

*Figure 1: Basic Building Diagram of the Proposed DBN-WP.*

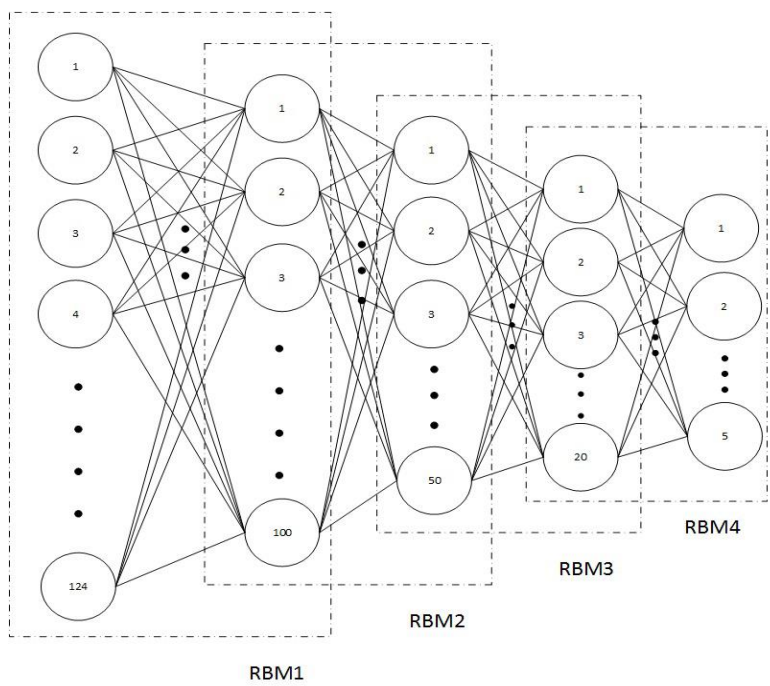

*Figure 2: Basic Architecture of DBN*

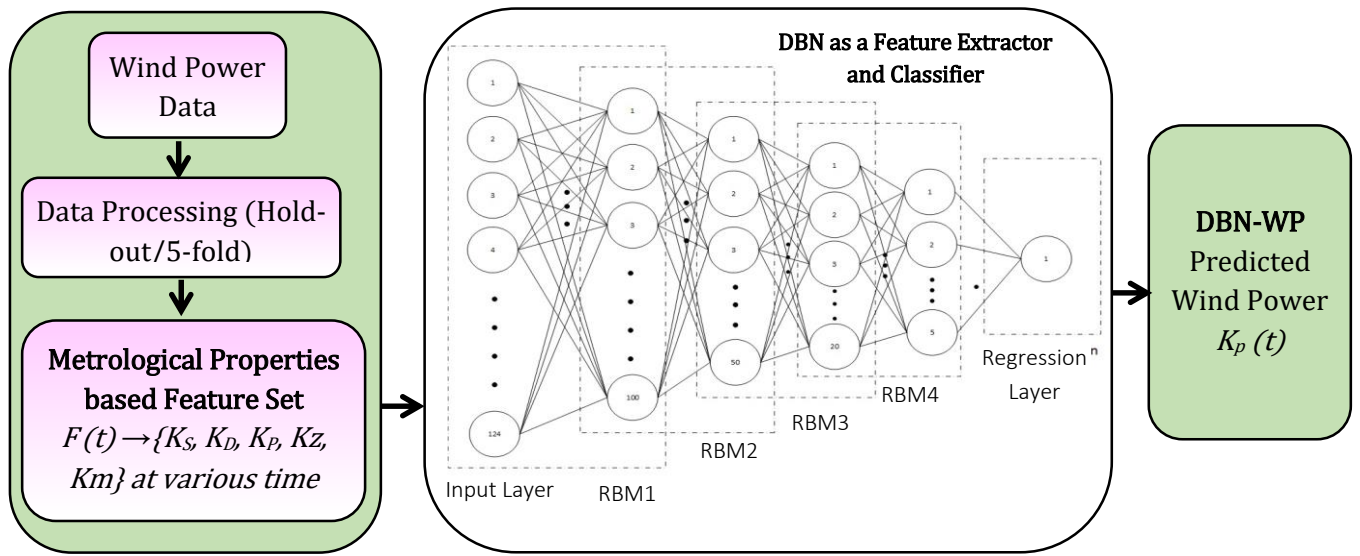

*Figure 3(a): Schematic Diagram of the Proposed DBN based Regressor for Wind Power Prediction.*

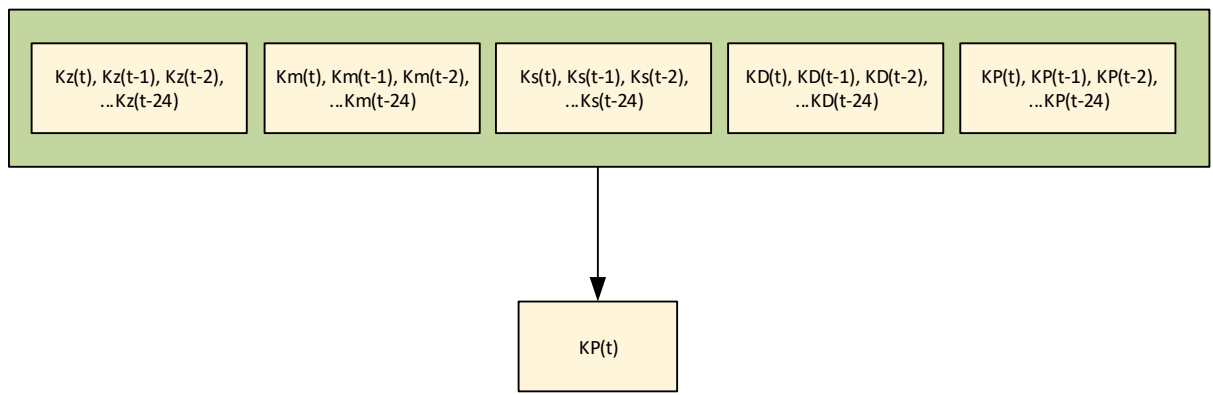

*Figure 3 (b): Dataset depicting input parameters of Feature set leading to output Wind Power Kp (t) at current time t.*

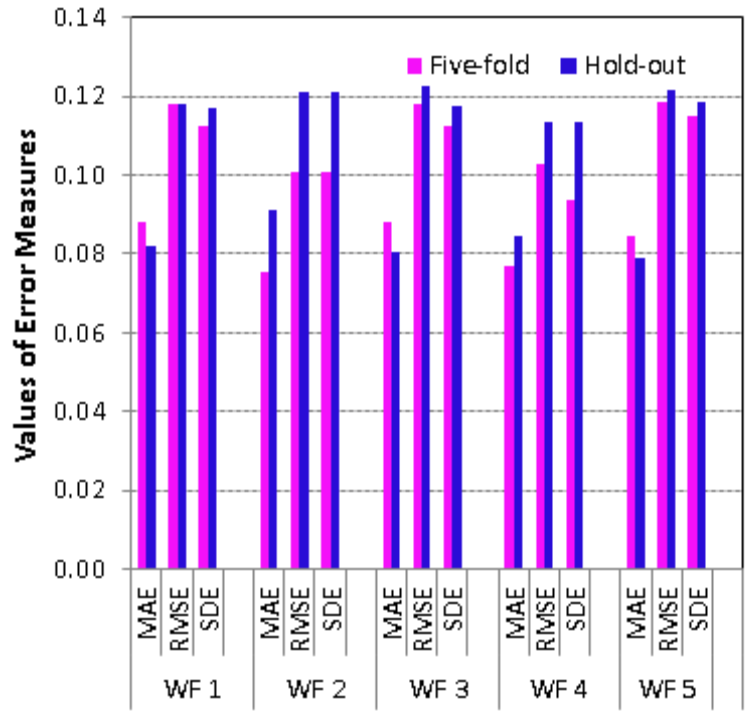

*Figure 4: Illustration of Error Measures of Wind Power Trained using Hold-out and 5-fold Data.*

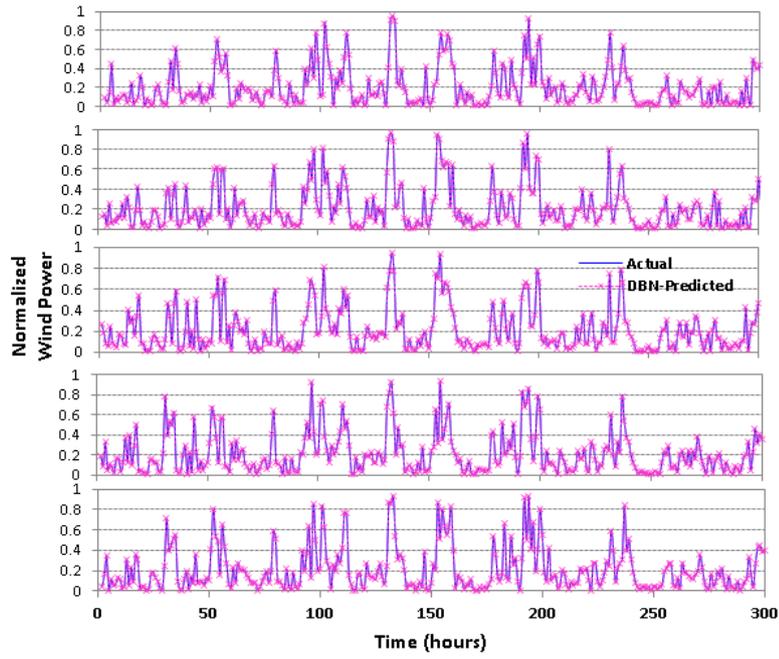

*Figure 5: Actual and DBN-Predicted Wind Power for Wind Farm 1 using 5-fold Test (from top to bottom).*

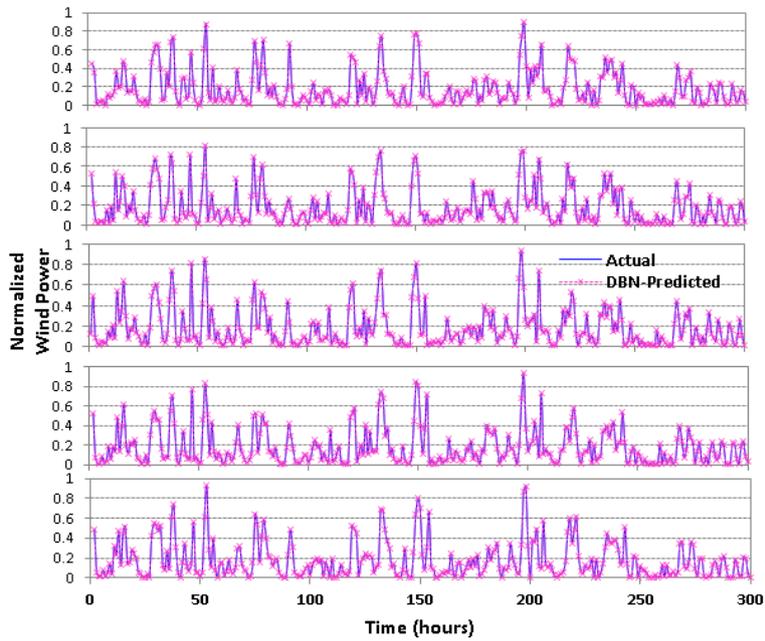

*Figure 6: Actual and DBN-Predicted Wind Power for Wind Farm 2 using 5-fold Test.*

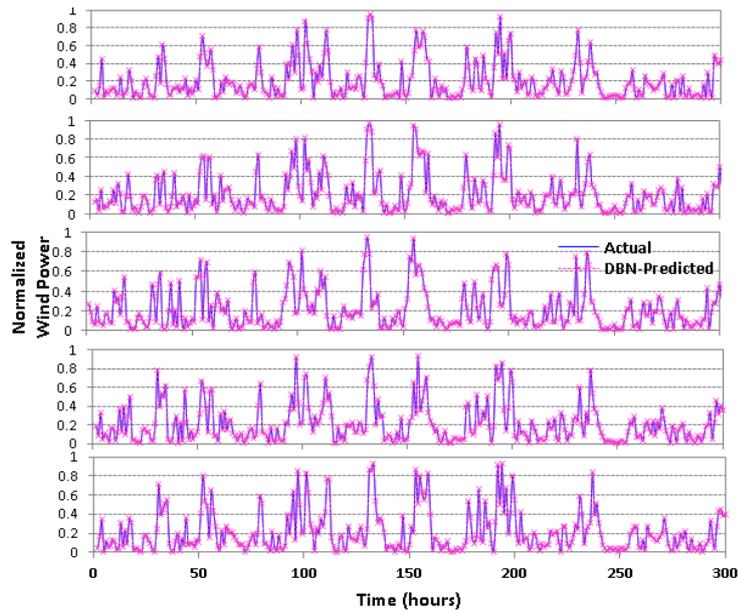

*Figure 7: Actual and DBN-Predicted Wind Power for Wind Farm 3 using 5-fold Test*

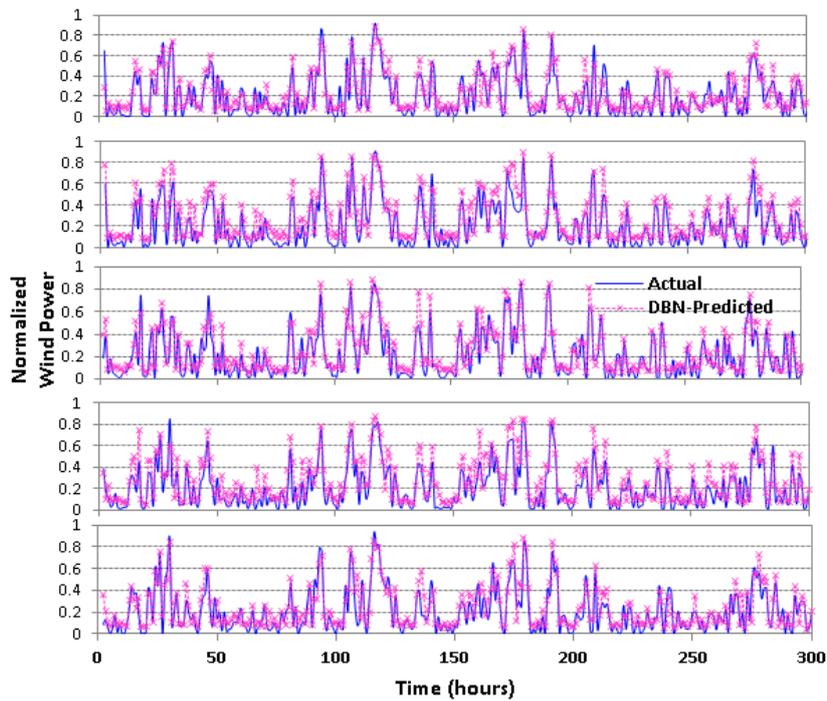

*Figure 8: Actual and DBN-Predicted Wind Power for Wind Farm 4 using 5-fold Test.*

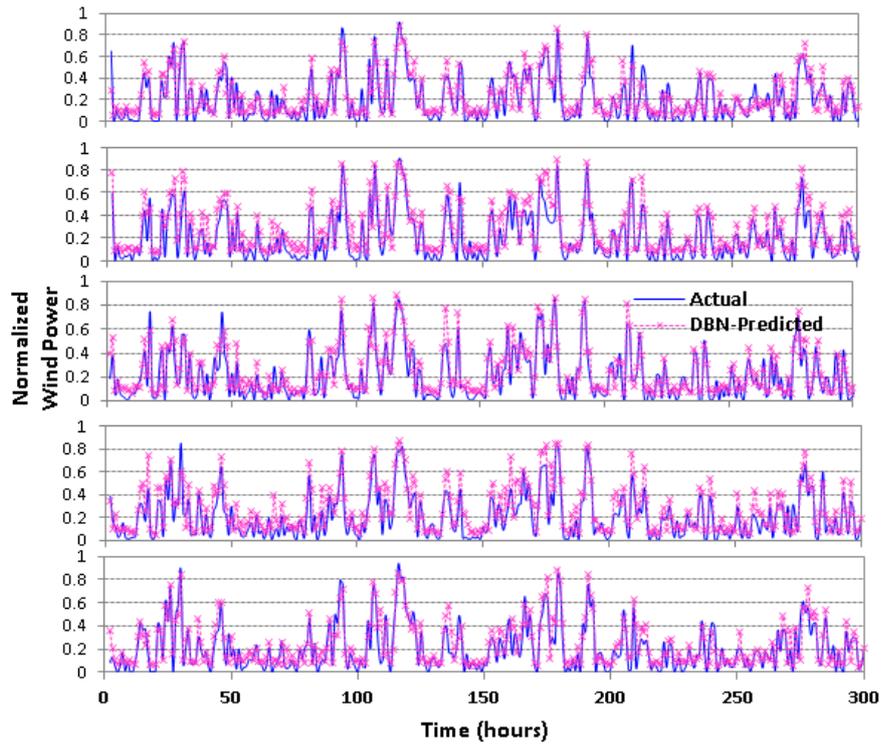

Figure 9: *Actual and DBN-Predicted Wind Power for Wind Farm 5 using 5-fold Test.*

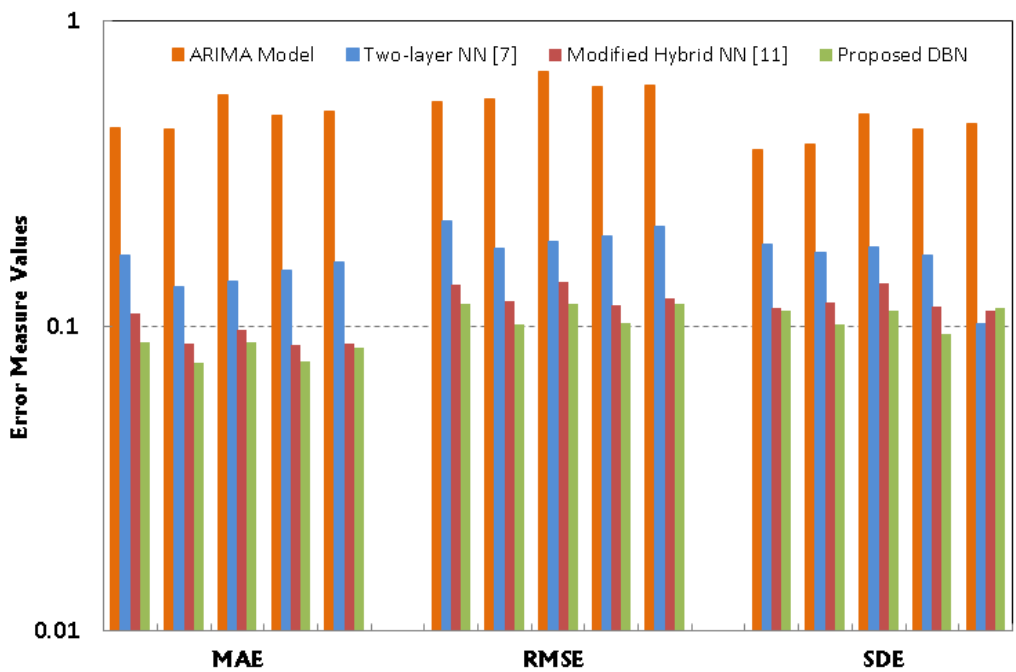

Figure 10: *Comparison of various techniques based on error measure values*

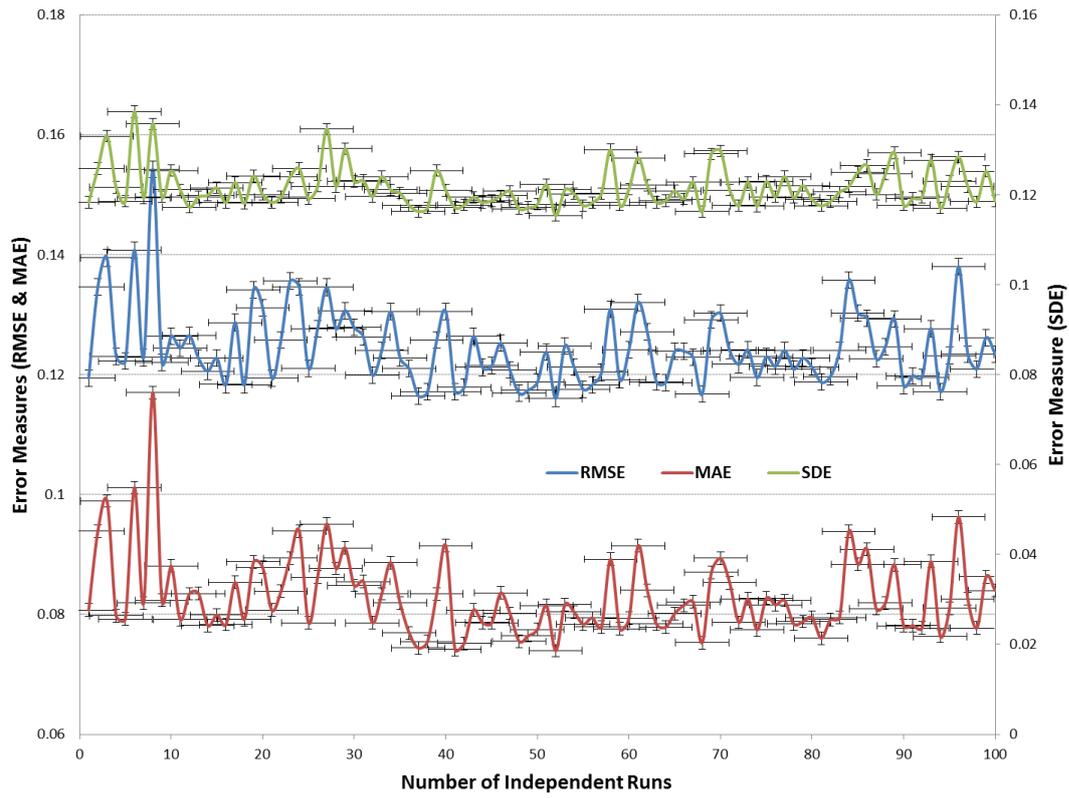

*Figure 11: Statistical Analysis for Hold-out Data Processing.*